\pdfoutput=1

\documentclass[11pt]{article}

\usepackage{acl}

\usepackage{multirow}
\usepackage{times}
\usepackage{adjustbox}
\usepackage{amssymb}
\usepackage{amsmath} 
\usepackage{latexsym}
\usepackage{times}
\usepackage{latexsym}
\usepackage{graphicx}
\usepackage{amssymb}
\usepackage[T1]{fontenc}

\usepackage[utf8]{inputenc}

\usepackage{microtype}

\usepackage{inconsolata}

\usepackage{graphicx}

\title{Sumotosima: A Framework and Dataset for Classifying and Summarizing Otoscopic Images}

\author{Eram Anwarul Khan$^{\dagger}$\thanks{ The authors are jointly the first authors}, Anas Anwarul Haq Khan$^{\ddagger *}$\\ { }
$^{\dagger}$Bombay Hospital, Department of ENT, $^{\ddagger}$IIT Bombay, Department of CSE\\
\texttt{\{eramkhan1607, anas290816007}\}@gmail.com}

\begin{document}
\maketitle
\begin{abstract}
Otoscopy is a diagnostic procedure to examine the ear canal and eardrum using an otoscope. It identifies conditions like infections, foreign bodies, ear drum perforations and ear abnormalities. We propose a novel resource efficient deep learning and transformer based framework, \textbf{Sumotosima} (\textbf{Sum}marizer for \textbf{otos}copic \textbf{ima}ges), an end-to-end pipeline for classification followed by summarization. Our framework works on combination of triplet and cross-entropy losses. Additionally, we use Knowledge Enhanced Multimodal
BART whose input is fused textual and image embedding. The objective is to provide summaries that are well-suited for patients, ensuring clarity and efficiency in understanding otoscopic images. Given the lack of existing datasets, we have curated our own \textbf{OCASD} (\textbf{O}toscopic \textbf{C}lassification \textbf{A}nd \textbf{S}ummary \textbf{D}ataset), which includes 500 images with 5 unique categories annotated with their class and summaries by Otolaryngologists. Sumotosima achieved a result of \textbf{98.03}\%, which is \textbf{7.00}\%, \textbf{3.10}\%, \textbf{3.01}\% higher than K-Nearest Neighbors, Random Forest and Support Vector Machines, respectively, in classification tasks. For summarization, Sumotosima outperformed GPT-4o and LLaVA by \textbf{88.53}\% and \textbf{107.57}\% in ROUGE scores, respectively. We have made our code and dataset publicly available at \footnote{\url{https://github.com/anas2908/Sumotosima}}
\end{abstract}

\section{Introduction}
\label{sec:introduction}
Otoscopy is a medical procedure using an otoscope to visually inspect the ear canal and eardrum. It is performed by healthcare professionals such as otolaryngologists (Ear Nose Throat specialists). This examination helps identify conditions like ear infections, blockages, eardrum perforations, and other abnormalities. Otoscopy is crucial for detection in treatment of ear-related issues, ensuring better auditory health and preventing complications that can arise from untreated conditions. Our objective is to develop efficient summaries for otoscopic ear images that are both clear and well-defined for patients. A common complaint from patients is the lack of interaction and explanations from healthcare professionals, often due to the high workload and limited availability of medical staff, which limits the time doctors can spend with each patient. By improving the clarity of information through AI-generated summaries, we aim to enhance patient understanding and interaction with their healthcare providers. These summaries identify the disease if any and describe various conditions, including areas of redness, infection spots, ear drum perforations, etc. Further details are provided in Section \ref{sec:dataset}.To facilitate our research, we identified lack of availability in open-source otoscopic image datasets, and none of them accompanying summaries or captions. To address this, we curated our dataset comprising 500 images equally distributed into 5 categories : `Acute Otitis Media', `Cerumen Impaction', `Chronic Otitis Media', `Myringosclerosis' and `Normal' from sources such as \cite{POLAT2021,viscaino2020computer}, and publicly available images on Google. It was was then filtered to remove noisy and redundant data that could introduce bias into the supervised learning results. Additionally, the images were summarized by an otolaryngologist to ensure the dataset's robustness. Further details and Annotation guidelines are provided in Section \ref{sec:guidelines}. Our framework is divided into two stages: classification followed by summarry generation. We observed that separating the classification task by applying a combination of triplet loss \cite{schroff2015facenet} and cross-entropy loss and passing resultant information alongside the prompt to a Multimodal BART \cite{xing2021km} yielded superior results. The main contributions of our proposed research are as follows:
\begin{itemize}
    \item \textbf{(1)} To best of our knowledge, \textbf{Sumotosima} is the first work towards understanding and summarization of otoscopic images.
    \item \textbf{(2)} We have curated first and the largest multi-modal dataset till date, comprising 500 instances, belonging to 5 unique categories. Each instance in the dataset is accompanied by a Gold Standard summary created by Otolaryngologist itself.
    \item \textbf{(3)} In lieu of resource-intensive models, we propose Sumotosima, an approach that involves SoTA classification followed by a summary generation through text and image embedding fusion before passing it to Multimodal BART.
    \item \textbf{(4)} The proposed method achieved the best classification results of \textbf{98.03}\%, with an improvement of \textbf{3.01}\% compared to previously utilized traditional machine learning approaches. For the summarization task, it demonstrated an improvement of \textbf{88.53}\% compared to GPT-4o
\end{itemize}

\section{Related Works}
\label{sec:relatedworks}

The classification of otoscopic images using traditional machine learning architectures has been extensively studied. Notable and recent research by \cite{bacsaran2020otitis, goshtasbi2020machine, crowson2021machine} has demonstrated the potential of machine learning for diagnosing ear conditions using otoscopic images. In recent years, Convolutional Neural Networks (CNNs) have become a popular approach for analyzing otoscopic images. These networks have shown better accuracy in detecting common ear diseases such as otitis media, tympanic membrane perforations, and cerumen impaction, as highlighted by studies from \cite{wu2021deep, bacsaran2020convolutional, tsutsumi2021web}. Moreover, both Machine Learning (ML) and Deep Learning (DL) algorithms have been employed to classify various ear conditions and segment areas of interest in otoscopic images, aiding in more precise diagnosis and treatment planning, as demonstrated by \cite{pham2021ear, pham2021tympanic}. Additionally, the integration of Artificial Intelligence (AI) with smartphone-connected otoscopes has facilitated remote diagnosis by analyzing images captured using smartphone attachments, providing real-time feedback to healthcare providers and patients, as shown by \cite{cortes2024smartphone}. Despite significant advancements in classification and segmentation tasks of otoscopic images using AI, there remains a lack of open-source datasets in this field. Furthermore, no substantial work has been done on summarizing otoscopic findings for better patient understanding. Addressing these gaps could significantly enhance the application of AI in otoscopy, which is the focus of our current research.

\section{Dataset}
\label{sec:dataset}

\begin{table*}[h]
\centering
\adjustbox{width=\textwidth}{
\begin{tabular}{|l|c|c|c|c|c|}
\hline
Dataset & Availability & Total Images & Unique Images & Classes & Annotated Summary \\\hline
\cite{dubois2024development}  & Closed Source   & 45,606      & -   & 11      & \texttimes            \\\hline
\cite{POLAT2021}       & Open Source             & 282     & \checkmark  & 7      & \texttimes             \\\hline
\cite{viscaino2020computer}  & Open Source & 880  & \checkmark    & 4      & \texttimes             \\\hline
OCASD  & Open Source & 500  & \checkmark            & 5      & \checkmark         \\\hline  
\end{tabular}}
\caption{Comparison of otoscopic image datasets based on their availability, total number of images, unique images, number of classes, and whether they include annotated summaries.}
\label{tab:comparison}
\end{table*}

\begin{table*}[h]
\adjustbox{width=\textwidth}{
\begin{tabular}{|l|c|c|c|c|}
\hline
Categories           & Source            & Intial Instances & Instances after Filtering & Instances Included \\ \hline
CI                   & \multirow{4}{*}{\cite{viscaino2020computer}} & 220              & 134                       & 100                \\ \cline{1-1} \cline{3-5} 
COM                  &                   & 220              & 174                       & 100                \\ \cline{1-1} \cline{3-5} 
MS                   &                   & 220              & 141                       & 100                \\ \cline{1-1} \cline{3-5} 
N                    &                   & 220              & 166                       & 100                \\ \hline
\multirow{2}{*}{AOM} &  \cite{POLAT2021} & 83               & 74                        & 69                 \\ \cline{2-5} 
                     & Internet          & 147              & 42                        & 31                 \\ \hline
\end{tabular}}
\caption{Dataset filtering process showing the initial instances, instances after filtering, and the final instances included for each category from various sources. Categories include Cerumen Impaction (CI), Chronic Otitis Media (COM), Myringosclerosis (MS), Normal (N), and Acute Otitis Media (AOM).}
\label{tab:filtering}
\end{table*}

Our dataset, \textbf{OCASD} (\textbf{O}toscopic \textbf{C}lassification \textbf{A}nd \textbf{S}ummary \textbf{D}ataset), was curated in response to the scarcity of available datasets and is derived from previously existing open-source datasets \cite{POLAT2021,viscaino2020computer}. Instances in \cite{viscaino2020computer} were found to be extensively redundant for the categories `Cerumen Impaction,' `Chronic Otitis Media,' `Myringosclerosis,' and `Normal.' To address this, we manually removed redundant or nearly identical instances, and an Otolaryngologist handpicked 100 images for each of the four categories, ensuring they were the most informative and contributive. A similar approach was followed for \cite{POLAT2021} for category `Acute Otitis Media' however, due to a lack of informative images, some were additionally sourced from the internet. The distribution of the curated collection is shown in Table \ref{tab:filtering}. Another relatively large dataset \cite{dubois2024development}, containing 11 classes and more than 45,000 images, is not accessible (refer Table\ref{tab:comparision}). After curation, OCASD comprises a total of 500 images across 5 unique categories. Each image is annotated with a summary by an Otolaryngologist, following the steps outlined in Section \ref{sec:guidelines}
\subsection{Annotation Steps and Guidelines}
\label{sec:guidelines}
\textbf{Objective} : To create clear, concise, and patient-friendly summaries of otoscopic images, identifying the disease (if any) and describing various conditions such as redness, ear wax, infection spots, etc.

\subsubsection{\textbf{General Guidelines}}

\textbf{Clarity and Simplicity}
\begin{itemize}
    \item Use simple and non-technical language that patients can easily understand.
    \item Avoid medical jargon; if medical terms must be used, provide a brief explanation.
\end{itemize}

\textbf{Empathy and Reassurance}
\begin{itemize}
    \item Write in a reassuring and empathetic tone.
    \item Aim to alleviate any anxiety by explaining conditions in a calm and positive manner.
\end{itemize}

\textbf{Consistency}
\begin{itemize}
    \item Follow a consistent structure for all summaries to ensure uniformity and ease of understanding.
\end{itemize}

\textbf{Accuracy}
\begin{itemize}
    \item Ensure that all information is accurate and reflects the observed condition.
\end{itemize}

\subsubsection{\textbf{Steps for Creating Summaries}}

\textbf{Identify the Category}
\begin{itemize}
    \item Confirm the known category of the image (e.g., Cerumen Impaction, Chronic Otitis Media, Myringosclerosis, Normal, Acute Otitis Media).
\end{itemize}

\textbf{Observe and Describe Key Features}
\begin{itemize}
    \item Note any visible signs of disease or abnormalities such as redness, swelling, infection spots, foreign body, wax buildup, perforations, or other notable conditions.
    \item Describe these features in simple terms.
\end{itemize}

\textbf{Summarize the Condition}
\begin{itemize}
    \item Begin with a brief statement identifying the disease (if any) or stating that the ear appears normal.
    \item Example: "This Otoscopic image shows signs of Chronic Otitis Media, which is an infection of the middle ear."
\end{itemize}

\textbf{Explain Symptoms and Impact}
\begin{itemize}
    \item Describe how the observed condition might affect the patient, focusing on symptoms they might experience.
    \item Example: "You may experience symptoms like ear pain, hearing loss, or discharge from the ear."
\end{itemize}

Review the summary for clarity, accuracy, and tone.
Make revisions as necessary to ensure the summary is patient-friendly and informative.

\begin{figure*}
  \includegraphics[width=\textwidth]{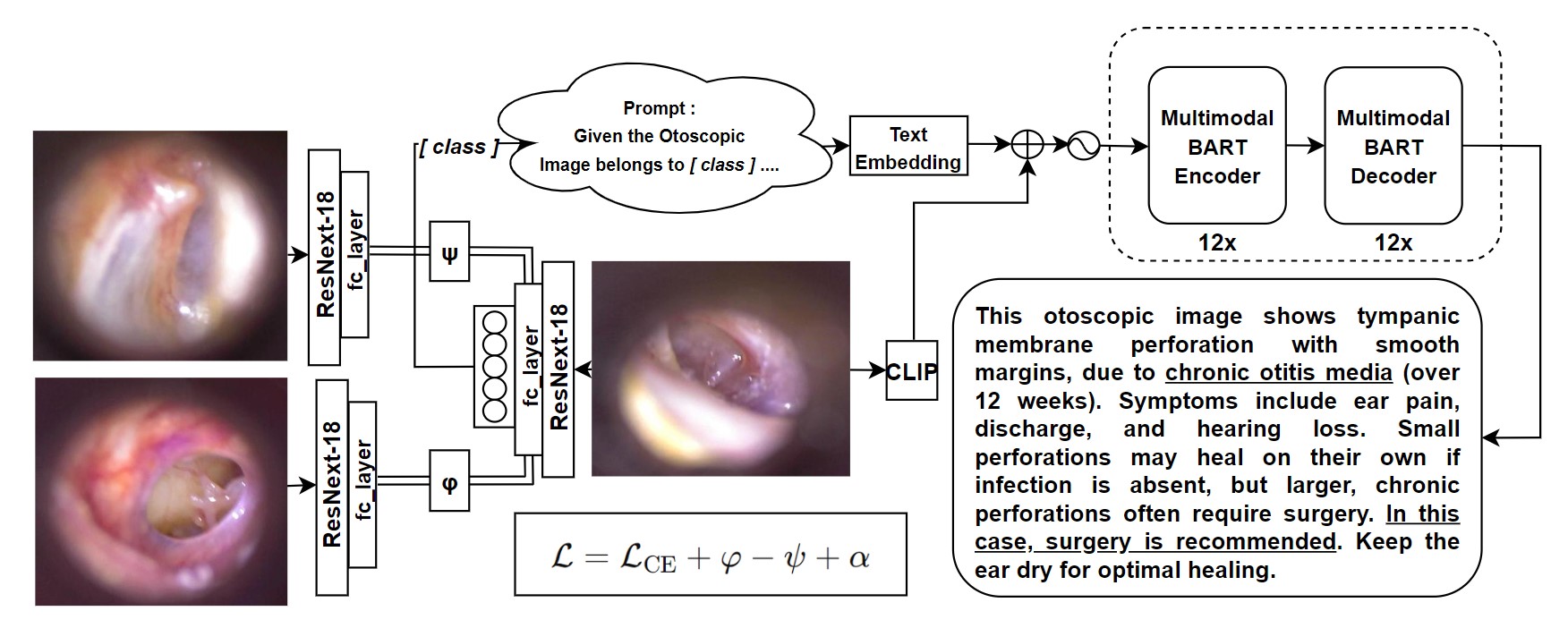}
  \caption{Overview of the end-to-end framework, \textbf{Sumotosima} (\textbf{Sum}marizer for \textbf{otos}copic \textbf{ima}ges), designed for classifying and generating summaries of otoscopic images.}

  \label{fig:model}
\end{figure*}

\section{Methodology}
\label{sec:methodology}
\subsection{Classification}

We followed a two-step architecture for our classification task. First, for each image (Anchor), we selected two additional images: one from the same category (Positive) and one from a different category (Negative). This setup facilitates use of triplet loss \cite{schroff2015facenet} alongside cross-entropy loss for classification.

The images were resized to \(226 \times 226\) and passed through ResNeXt-18 \cite{he2016deep} to obtain vector features of \(\mathbb{R}^{\scriptsize{1 \times 128}}_{\text{anc}}\), \(\mathbb{R}^{\scriptsize{1 \times 128}}_{\text{pos}}\) and \(\mathbb{R}^{\scriptsize{1 \times 128}}_{\text{neg}}\) for Anchor, Positive, and Negative images respectively. The Euclidean distance between \(\mathbb{R}^{\scriptsize{1 \times 128}}_{\text{anc}}\) and \(\mathbb{R}^{\scriptsize{1 \times 128}}_{\text{pos}}\) is denoted by \(\varphi\), and the distance between \(\mathbb{R}^{\scriptsize{1 \times 128}}_{\text{anc}}\) and \(\mathbb{R}^{\scriptsize{1 \times 128}}_{\text{neg}}\) is denoted by \(\psi\). The objective is to minimize \(\varphi\) and maximize \(\psi\), with a margin \(\alpha\) to enforce a minimum distance between positive and negative pairs, thus enhancing the discriminative capability of the learned embeddings. \(\alpha\) is set to the default value \cite{schroff2015facenet}.

\begin{equation}
a_i = \mathbb{R}^{\scriptsize{1 \times 128}}_{\text{anc}}, \quad p_i = \mathbb{R}^{\scriptsize{1 \times 128}}_{\text{pos}}, \quad n_i = \mathbb{R}^{\scriptsize{1 \times 128}}_{\text{neg}}
\end{equation}

\begin{equation}
\varphi = \| a_i - p_i \|_2^2, \quad \psi = \| a_i - n_i \|_2^2
\end{equation}

The triplet loss is given by:
\begin{equation}
L_{\text{triplet}} =  \sum_{i=1}^N \left[ \varphi - \psi + \alpha \right]_+
\end{equation}

The cross-entropy loss is defined as:
\begin{equation}
L_{\text{CE}} = -\frac{1}{N} \sum_{i=1}^{N} \sum_{c=1}^{C} y_{i,c} \log(\hat{y}_{i,c})
\end{equation}

\begin{itemize}
    \item \(N\) is the total number of image instances.
    \item \(C\) is the number of classes (set to 5).
    \item \(y_{i,c}\) is the binary indicator (0 or 1) if class label \(c\) is the correct classification for instance \(i\).
    \item \(\hat{y}_{i,c}\) is the predicted probability of instance \(i\) being class \(c\).
\end{itemize}

The total loss while training the model is:
\begin{equation}
L = L_{\text{triplet}} + L_{\text{CE}}
\end{equation}

The loss backpropagates and adjusts the fully connected layer of the model, denoted as \textit{fc\_layer} in Figure \ref{fig:model}, thereby refining the vector representation of size \(\mathbb{R}^{\scriptsize{1 \times 128}}\) in each epoch. After classification, the class is included as part of the prompt for robust results.

\subsection{Generation}
The anchor image undergoes preprocessing through transformations, including resizing, center cropping, and normalization. The processed image is then fed into a fine-tuned CLIP~\cite{radford2021learning}, yielding a dense vector \(\mathbf{v}_{\text{CLIP}} \in \mathbb{R}^{1 \times 512}\). The resulting embeddings \(\mathbf{v}_{\text{CLIP}}\) are combined with the text embeddings \(\mathbf{e}_{\text{prompt}} \in \mathbb{R}^{d}\), which are obtained by passing the class-enriched prompt \(\mathbf{p}_c\) through the KM-BART encoder. Although CLIP has not been extensively trained on medical image datasets, it has effectively learned spatial, textual, and color characteristics of images. After fine-tuning on OCASD, CLIP can identify fine details such as the granularity of middle ear parts, the level of redness, potential irritability, and the shape of tympanic membranes. This capability is crucial for determining the criticality of the image in relation to the class specified by the prompt. The fused embeddings \(\mathbf{v}_{\text{fused}} = \mathbf{v}_{\text{CLIP}} + \mathbf{e}_{\text{prompt}}\) are then augmented with positional encodings as described in~\cite{vaswani2017attention}, before being passed to the subsequent KM-BART encoder. Let \(k\) denote the position index and \(i\) the dimension index. The updated positional encoding is given by:

\begin{equation}
\mathbf{v}'_{\text{fused}} = \mathbf{v}_{\text{fused}} + \mathbf{r}_{k/i},
\end{equation}

where \(\mathbf{r}_{k/i}\) represents the positional encoding adjustment based on the relative position \(k\) and dimension \(i\).

\section{Evaluation Metrics}
\label{sec:metrics}
For Classification task we employed Precision, Recall and F1-score. \\
\textbf{Precision}: The proportion of true positive predictions among all positive predictions. It measures the accuracy of positive predictions.
\[
\text{Precision} = \frac{\text{True Positives}}{\text{True Positives} + \text{False Positives}}
\]

\textbf{Recall}: The proportion of true positive predictions among all actual positives. It measures how well the model identifies all relevant cases.
\[
\text{Recall} = \frac{\text{True Positives}}{\text{True Positives} + \text{False Negatives}}
\]

\textbf{F1-score}: The harmonic mean of Precision and Recall. It provides a single metric that balances both precision and recall.
\[
\text{F1-score} = 2 \cdot \frac{\text{Precision} \cdot \text{Recall}}{\text{Precision} + \text{Recall}}
\]

For Generated Summary, we evaluated our model using Automatic metrics and Human Evaluation. For Automatic Evaluation, we used BLEU~\cite{papineni2002bleu} for translation quality, BERTScore~\cite{zhang2019bertscore} for fluency and coherence and ROUGE~\cite{lin2004rouge} for summarization quality. Human Evaluation included Patient Friendliness and Faithfulness. Faithfulness was annotated by our in-house Otolaryngologist as the percentage of error-free samples. For assessing Patient Friendliness, we employed seven annotators with diverse backgrounds: three with some knowledge of medical terminology from science background, and four working professionals—two men and two women—from commerce and arts. Each annotator rated 100 summaries on a three-point scale: \textbf{1 for "no use,"} indicating the summary provided no valuable information; \textbf{2 for "somewhat useful,"} meaning it included general information about the disease class, such as symptoms like redness, irritability, or ear pain, along with trivial suggestions; and \textbf{3 for "very useful,"} where the summary addressed the specific scenario, detailing whether surgery is required or identifying the exact part of the ear affected, offering non-trivial insights that require image analysis. This approach ensured a comprehensive evaluation of patient friendliness from both semi-informed and layperson perspectives.

\section{Experiments and Results}
\label{sec:experiments}
\subsection{Classification}
For classification, we compared machine learning approaches such as KNN \cite{guo2003knn}, Random Forest \cite{breiman2001random}, and Support Vector Machine. Support Vector Machine \cite{cortes1995support} was found to perform best among the machine learning algorithms. We then applied the ResNeXt-18 architecture for the deep learning classification task and experimented with cross-entropy and triplet loss functions. We observed that the combination of cross-entropy and triplet loss functions performed best. To handle any bias or error due to the scarcity of the dataset, we performed a \textbf{5-fold cross-validation} approach. The average results observed showed a \textbf{3.15\%} gain over the best performing machine learning approach, a \textbf{1.34\%} gain over the deep learning architecture with a single cross-entropy loss, and a \textbf{1.55\%} gain over the deep learning architecture with a single triplet loss. Further refer Table \ref{tab:classificationresult}


\begin{table}[h]
\scalebox{0.87}{
\begin{tabular}{l|ll|l}
\hline
Models                                  & P     & R     & F1    \\ \hline
K Nearest Neighbour                    & 0.911 & 0.910 & 0.910 \\
Random Forest                           & 0.954 & 0.950 & 0.951 \\
Support Vector Machine                  & 0.953 & 0.950 & 0.950 \\ \hline
ResNeXt18 ($L_{\text{CE}}$)                      & 0.971 & 0.967 & 0.968 \\
ResNeXt18 ($L_{\text{trp}}$)                     & 0.954 & 0.978 & 0.966 \\ \hline
ResNeXt18 ($L_{\text{CE + trp}}^{\text{fold 1}}$) & 0.973 & 0.963 & 0.965 \\
ResNeXt18 ($L_{\text{CE + trp}}^{\text{fold 2}}$) & 0.976 & 0.975 & 0.975 \\
ResNeXt18 ($L_{\text{CE + trp}}^{\text{fold 3}}$) & 1     & 1     & 1     \\
ResNeXt18 ($L_{\text{CE + trp}}^{\text{fold 4}}$) & 0.988 & 0.988 & 0.988 \\
ResNeXt18 ($L_{\text{CE + trp}}^{\text{fold 5}}$) & 0.978 & 0.975 & 0.975 \\ \hline
\textbf{ResNeXt18} ($L_{\text{CE + trp}}^{\text{avg}}$) & \textbf{0.983} & \textbf{0.980} & \textbf{0.981} \\ \hline
\end{tabular}}
\caption{Performance comparison of different models on the classification task using Precision ($P$), Recall ($R$), and F1-score ($F1$). The best results are achieved by ResNeXt18 using a combination of cross-entropy and triplet loss, with the average results across five folds highlighted. The z-score is approximately 3.81, and the two-tailed p-value is approximately 0.00014. This very small p-value suggests that the difference between the proportions 0.983 (ours) and 0.953 (SVM) is statistically significant.}
\label{tab:classificationresult}
\end{table}

\subsection{Generation}

For generation, we compared our model with the open-source vision-language model LLaVA \cite{liu2023improvedllava} and the closed-source GPT-4o with the given prompt : \textit{Given an otoscopic image, generate a summary ensuring clarity and simplicity by using simple and non-technical language that patients can easily understand, avoiding medical jargon unless a brief explanation is provided. When creating summaries, first identify the category of the image (e.g., Cerumen Impaction, Chronic Otitis Media, Myringosclerosis, Normal, Acute Otitis Media). Observe and describe any visible signs of disease or abnormalities such as redness, swelling, infection spots, wax buildup, perforations, or other notable conditions in simple terms. Begin the summary with a brief statement identifying the disease (if any) or stating that the ear appears normal. For example, "This otoscopic image shows signs of Chronic Otitis Media, which is an infection of the middle ear." Explain how the observed condition might affect the patient, focusing on symptoms they might experience, such as, "You may experience symptoms like ear pain, hearing loss, or discharge from the ear." Finally, review the summary for clarity, accuracy, and tone, making revisions as necessary to ensure it is patient-friendly and informative. Everything must be in one paragraph.} For the prompt to Sumotosima, we already predefined the class found from the classification model. We also experimented with the following variations of Sumotosima:

\begin{itemize}
    \item \textbf{$Sumotosima_{\text{1}}$}: This variant uses \underline{traditional BART} \cite{lewis2019bart} and not KM-BART. The image information is first converted to \underline{captions} using a \underline{pretrained CLIP} model and is passed as input along with the prompt.
    \item \textbf{$Sumotosima_{\text{2}}$}: This variant uses \underline{traditional BART} \cite{lewis2019bart} and not KM-BART. The image information is extracted as \underline{dense vectors} using a \underline{pretrained CLIP} model and is passed as input along with the prompt.
    \item \textbf{$Sumotosima_{\text{3}}$}: This variant uses \underline{traditional BART} \cite{lewis2019bart} and not KM-BART. The image information is first converted to \underline{captions} using a \underline{Finetuned CLIP} model on OCASD.
    \item \textbf{$Sumotosima_{\text{4}}$}: This variant uses \underline{traditional BART} \cite{lewis2019bart} and not KM-BART. The image information is extracted as \underline{dense vectors} using a \underline{Finetuned CLIP} model on OCASD.
    \item \textbf{$Sumotosima_{\text{5}}$}: This variant uses \underline{KM-BART}. The image information is extracted as \underline{dense vectors} using a \underline{Finetuned CLIP} model on OCASD.
\end{itemize}

We found $Sumotosima_{\text{5}}$ to give the best results among all the ablation studies and vision-language models used as baselines. Specifically, $Sumotosima_{\text{5}}$ surpassed GPT-4o by \textbf{88.53\%} and LLaVA by \textbf{107.57\%} in ROUGE scores. Given that Sumotosima is fine-tuned, it may produce summaries that perform better on automatic evaluation metrics like BLEU and ROUGE. This questions the reliability of these scores. To address this issue, we incorporated BERT-F1 as part of our evaluation framework. $Sumotosima_{\text{5}}$ also surpassed GPT-4o by \textbf{17.65\%} and LLaVA by \textbf{23.09\%} in BERT-F1 scores. Further refer Table \ref{tab:generationresult}


\begin{table}[h]
\scalebox{0.9}{
\begin{tabular}{l|lll}
\hline
Model                      & BLEU   & BERT-F1 & ROUGE-L \\ \hline
GPT-4o                     & 0.0348 & 0.442   & 0.185   \\
LLaVA                      & 0.029  & 0.423   & 0.168   \\ \hline
$Sumotosima_{\text{1}}$    & 0.063  & 0.285   & 0.228   \\
$Sumotosima_{\text{2}}$    & 0.071  & 0.296   & 0.184   \\
$Sumotosima_{\text{3}}$    & 0.092  & 0.318   & 0.253   \\
$Sumotosima_{\text{4}}$    & 0.101  & 0.347   & 0.281   \\
$Sumotosima_{\text{5}}$ & \textbf{0.128} & \textbf{0.521} & \textbf{0.349} \\ \hline
\end{tabular}}
\caption{Comparison of different models for summarization tasks using BLEU, BERT-F1, and ROUGE-L metrics. $Sumotosima_{\text{5}}$ achieved the best performance across all metrics. The Z-score is approximately $8.29$, and the two-tailed p-value is approximately $2.22 \times 10^{-16}$. This extremely small p-value indicates a statistically significant difference between GPT-4o and $Sumotosima_{\text{5}}$.}
\label{tab:generationresult}
\end{table}

We also conducted human evaluation as mentioned in Section \ref{sec:metrics}. We found $Sumotosima_{\text{5}}$ to be \textbf{19\%} more faithful in the generation of summaries and \textbf{33.33\%} more patient-friendly than GPT-4o. Among all Human Evaluation Metrics, LLaVA was the worst performing. Further refer Table \ref{tab:humanevaluation}

\begin{table}[h]
\scalebox{0.9}{
\begin{tabular}{l|ll}
\hline
Model         & Patient Friendliness & Faithfulness \\ \hline
GPT-4o        & 1.8                  & 73.0\%       \\
LLaVA         & 1.2                  & 38.5\%       \\ 
$Sumotosima_{\text{5}}$ & \textbf{2.4}                  & \textbf{92.0\%}       \\ \hline
\end{tabular}}
\caption{Human Evaluation of models on patient friendliness and faithfulness. Sumotosima\textsubscript{5} outperformed GPT-4o and LLaVA in both criteria.}
\label{tab:humanevaluation}
\end{table}

\subsection{Experimental Setup}
We performed a grid search for hyperparameters using the Adam optimizer and an Nvidia A100 GPU for a total of 3 hours. Refer to Table \ref{tab:parameters}

\begin{table}[h]
\scalebox{0.9}{
\begin{tabular}{|c|c|c|}
\hline
Parameters/Resources &Classification & Generation \\ \hline
Training Split         & 70\%                        & 60\%                    \\
Validation Split       & 15\%                        & 20\%                    \\
Test Split             & 15\%                        & 20\%                    \\
Epochs                 & 100                         & 50                      \\
Batch Size             & 32                          & 8                       \\
Learning Rate             & 1e-3                          & 3e-5                       \\
GPU                    & 4.5GB      & 18GB     \\ \hline
\end{tabular}}
\caption{Training parameters and resource allocation for classification and generation tasks, including dataset splits, epochs, batch size, learning rate, and GPU usage.}
\label{tab:parameters}
\end{table}

\section{Conclusion}
In this work, we introduced a novel pipeline for classifying and generating summaries for otoscopic images of the middle ear, with the objective of developing summaries that are both well-defined and patient-friendly, addressing the challenge of insufficient explanations from medical professionals due to their hectic schedules and limited time per patient. Unlike previous approaches that relied on traditional machine learning algorithms or straightforward deep learning architectures, our model utilizes a combination of triplet loss and cross-entropy loss, built upon the ResNeXt-18 architecture, achieving a classification accuracy of \textbf{98.03\%}, surpassing all baselines on our OCASD dataset, which comprises 500 images across 5 different classes. As this is the first work on summarization for otoscopic images, we addressed the lack of annotated summary datasets by creating summaries for all 500 images, combining class-enriched prompts with image embeddings obtained from a fine-tuned CLIP model, and feeding these into a Multimodal BART model for summary generation. Our framework, Sumotosima, significantly outperformed GPT-4o and LLaVA in summarization, with improvements of \textbf{88.53\%} and \textbf{107.57\%} in ROUGE scores, respectively. Looking ahead, we plan to extend this work by incorporating additional metadata, such as patient age, gender, medical history, and educational background, to generate more robust and patient-specific summaries.

\bibliography{custom}
\bibliographystyle{acl_natbib}

\end{document}